\documentclass[sigconf,nonacm]{acmart}
\setcopyright{none}
\acmConference[WWW '27]{The ACM Web Conference 2027}{May 10--14, 2027}{Dublin, Ireland}
\acmYear{2027}
\usepackage{booktabs}
\usepackage{multirow}
\usepackage{xspace}
\usepackage{amsmath}
\usepackage{graphicx}
\usepackage{subcaption}
\usepackage{xcolor}
\xspaceaddexceptions{\%}

\newcommand{\sys}{ZeroPolar\xspace}
\newcommand{\sysT}{ZeroPolar-T\xspace}
\newcommand{\polarity}{structural polarity\xspace}

\newcommand{\nAllT}{73.99\xspace}
\newcommand{\nAllZ}{72.34\xspace}

\newcommand{\nAllBP}{67.98\xspace}

\newcommand{\nRegHomT}{71.03\xspace}
\newcommand{\nRegWebT}{79.18\xspace}

\newcommand{\nNbaseW}{eighteen\xspace}
\newcommand{\nDiffSage}{2.06\xspace}

\newcommand{\nDiffBP}{6.01\xspace}

\newcommand{\nAllLLM}{59.60\xspace}

\newcommand{\nPubBest}{47.98\xspace}
\newcommand{\nHetQwenBest}{64.40\xspace}

\newcommand{\nFreshDiff}{1.25\xspace}
\newcommand{\nLocleAdaFo}{6.63\xspace}
\newcommand{\nLocleAdaQw}{6.21\xspace}

\newcommand{\nGofaFresh}{19.05\xspace}

\newcommand{\nNbaseFreshW}{fourteen\xspace}
\newcommand{\nPatThirty}{73.42\xspace}
\newcommand{\nSeedFresh}{2.74\xspace}

\newcommand{\nSeedAll}{2.07\xspace}

\newcommand{\nBudPct}{5\xspace}
\newcommand{\nBudZ}{67.85\xspace}
\newcommand{\nBudSage}{57.05\xspace}

\newcommand{\nBudGap}{10.80\xspace}
\newcommand{\nFewK}{5\xspace}
\newcommand{\nFewKZ}{62.98\xspace}
\newcommand{\nFewKSage}{41.55\xspace}

\newcommand{\nCrossT}{71.30\xspace}

\newcommand{\nEnvCross}{68.70\xspace}
\newcommand{\nCtrlPlusT}{5.21\xspace}
\newcommand{\nCtrlMinusT}{24.54\xspace}
\newcommand{\nCtrlNoStepT}{1.25\xspace}
\newcommand{\nCtrlNoGateZ}{5.66\xspace}
\newcommand{\nCtrlNoPropZ}{1.34\xspace}
\newcommand{\nCtrlNoLabelZ}{4.59\xspace}
\newcommand{\nQwenT}{65.59\xspace}

\newcommand{\nQwenSage}{63.44\xspace}

\newcommand{\nNewT}{74.22\xspace}
\newcommand{\nLlamaZ}{63.52\xspace}
\newcommand{\nFsZten}{76.16\xspace}
\newcommand{\nFsBPten}{73.34\xspace}
\newcommand{\nSqTten}{65.36\xspace}
\newcommand{\nSqBPten}{68.07\xspace}

\newcommand{\nEncW}{seven\xspace}
\newcommand{\nEncTLo}{68.66\xspace}
\newcommand{\nEncTHi}{74.14\xspace}

\newcommand{\nEncTBestW}{six\xspace}
\newcommand{\nEncTBPMin}{5.66\xspace}
\newcommand{\nExtT}{54.12\xspace}
\newcommand{\nExtBP}{52.53\xspace}
\newcommand{\nDeltaQLo}{0.1\xspace}
\newcommand{\nSensZLo}{72.05\xspace}
\newcommand{\nSensZHi}{72.47\xspace}
\newcommand{\nNoiseMax}{40\xspace}
\newcommand{\nNoiseT}{68.55\xspace}

\newcommand{\nNoiseZthree}{62.95\xspace}
\newcommand{\nNoiseZone}{47.88\xspace}
\newcommand{\nNoiseLead}{15.07\xspace}
\newcommand{\nNoiseProtoLead}{14.65\xspace}
\newcommand{\nTimeZ}{0.10\xspace}
\newcommand{\nTimeSage}{12.91\xspace}
\newcommand{\nTimeT}{7.89\xspace}

\newcommand{\nMinusHomoMinAbs}{15.42\xspace}
\newcommand{\nPlusWebMinAbs}{9.27\xspace}
\newcommand{\nSHomoMin}{0.360\xspace}
\newcommand{\nSWebMax}{0.078\xspace}
\newcommand{\nDsHChild}{0.65\xspace}
\newcommand{\nTrueHChild}{0.422\xspace}

\newcommand{\nFixSport}{9,376\xspace}
\newcommand{\nBreakSport}{2,722\xspace}

\newcommand{\nFixWinsW}{nine\xspace}

\newcommand{\nFoneT}{65.01\xspace}

\begin{document}

\title{When Links Mislead: Label-Free Structural Polarity for Zero-Shot Classification of Web Pages}

\author{Hanshuai Cui}
\affiliation{\institution{School of Artificial Intelligence, Beijing Normal University, Beijing 100875, China}\city{}\country{}}
\affiliation{\institution{Institute of Artificial Intelligence and Future Networks, Beijing Normal University, Zhuhai 519087, China}\city{}\country{}}
\author{Zhiqing Tang}
\affiliation{\institution{Institute of Artificial Intelligence and Future Networks, Beijing Normal University, Zhuhai 519087, China}\city{}\country{}}
\author{Zhi Yao}
\affiliation{\institution{School of Artificial Intelligence, Beijing Normal University, Beijing 100875, China}\city{}\country{}}
\affiliation{\institution{Institute of Artificial Intelligence and Future Networks, Beijing Normal University, Zhuhai 519087, China}\city{}\country{}}
\author{Weijia Jia}
\affiliation{\institution{Institute of Artificial Intelligence and Future Networks, Beijing Normal University, Zhuhai 519087, China}\city{}\country{}}
\author{Jun Luo}
\affiliation{\institution{College of Computing and Data Science, Nanyang Technological University, Singapore}\city{}\country{}}

\begin{abstract}
Web platforms such as digital libraries, online stores and university sites must classify new pages before anyone labels them. Large language models (LLMs) can label each page from its text, but these labels are noisy, and links could correct them only where linked pages share classes.
Existing zero-shot methods ignore the links, or pull a page toward its neighbours as label propagation and graph neural networks (GNNs) do, which hurts where linked pages differ in class. Others weight propagation by an LLM-estimated homophily that costs extra calls and drifts across LLMs.
We find that the sign with which links should enter a prediction, which we call \emph{\polarity}, is readable without true labels. It is the excess agreement of LLM labels and text-prototype predictions along links.
We thus propose \sys, which gates one signed propagation step by this statistic.
Training-free \sys mixes prototype scores with any number of LLM labels, and \sysT denoises the labels with a text classifier once half of the pages carry one.
With one GPT-4o label per page, \sysT reaches the highest mean accuracy on eleven graphs (\nAllT\%), \nDiffBP points above LLM-BP and significantly above all \nNbaseW baselines we compare.
With LLM labels on only \nBudPct\% of the pages, training-free \sys leads a GNN trained on these labels by \nBudGap points.
\end{abstract}


\maketitle

\begin{figure}[t]
\centering
\includegraphics[width=\linewidth]{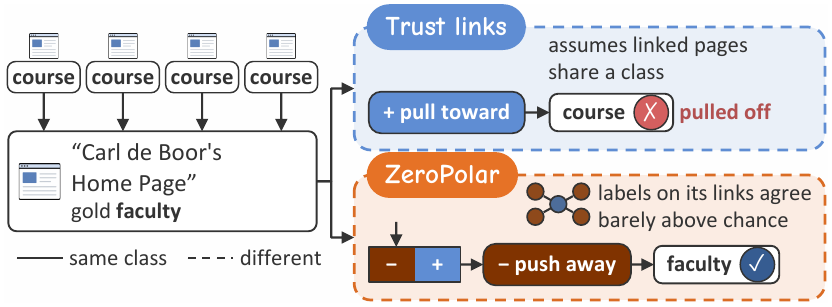}
\caption{\sys measures links instead of trusting them. A professor's page has four course in-links, so a step toward them gives course, while \sys finds that labels barely agree along the graph's links and pushes away.}
\Description{On the left, a professor's home page from the Wisconsin graph, whose true class is faculty, has four in-links from course pages, with a key for solid same-class and dashed different-class links. On the right, two stacked lanes. The upper lane, Trust links, assumes that linked pages share a class, pulls the page toward its neighbours and labels it course, marked wrong. The lower lane, \sys{}, reads that zero-shot labels along the graph's links agree barely above chance, sets the marker on the minus side of a minus and plus scale, pushes the page away from its neighbours and keeps faculty, marked correct.}
\label{fig:teaser}
\end{figure}

\section{Introduction}
\label{sec:intro}

New Web platforms, such as digital libraries, online stores, encyclopedias and university sites, publish pages and links long before anyone labels them.
Each page, whether a paper, a product, an encyclopedia entry or a page on a university department site, becomes a node of a text-attributed graph.
Zero-shot node classification assigns every page a category from its text and links alone, so that a platform can organize, search and moderate its content from the first day~\cite{chenexplore,llmgraphsurvey,llmbp}.
Large language models (LLMs) make this practical, because GPT-4o~\cite{gpt4o} or an open model such as Qwen3~\cite{qwen3} labels a page from its text in one call.
However, these labels are noisy, and the links that could correct them mean different things on different platforms.
A co-purchase link usually joins two products of the same category, while hyperlinks on a university site mostly join pages of different classes~\cite{webkb,geomgcn}, such as a professor's home page and the course pages that link to it (Figure~\ref{fig:teaser}).
Classifying Web pages from noisy LLM labels and such links is thus learning on noisy Web graphs, a topic of the Graph Algorithms and Modeling for the Web track.

Existing zero-shot methods use the links in one of three ways.
The first ignores them and takes the label of the nearest class prototype in a text embedding space or the label the LLM gives~\cite{bgeicl,llm2vec,llmnodebed}.
The second propagates labels along the links with a fixed rule~\cite{lp} or, like LLM-BP~\cite{llmbp}, runs belief propagation weighted by an edge homophily that an LLM estimates from sampled edges.
The third trains graph neural networks (GNNs)~\cite{gcn,graphsage} on LLM labels, as LLM-GNN does~\cite{llmgnn}.
Graph foundation models pretrained across graphs~\cite{ofa,zerog,uniglm,gofa} and LLMs tuned on graph tokens~\cite{graphgpt,llaga} also transfer zero-shot, yet on average they trail these simpler methods when the target graph is new~\cite{llmbp}.

\begin{figure}[t]
\centering
\begin{subfigure}[t]{120pt}\centering\includegraphics[width=120pt]{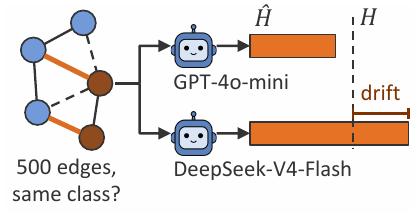}\caption{Homophily from LLM calls.}\end{subfigure}\hfill
\begin{subfigure}[t]{120pt}\centering\includegraphics[width=120pt]{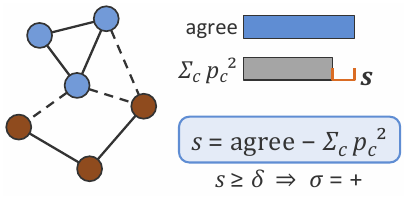}\caption{Excess agreement.}\end{subfigure}\\[2pt]
\begin{subfigure}[t]{115.5pt}\centering\includegraphics[width=115.5pt]{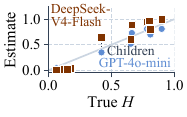}\caption{It drifts with the LLM.}\end{subfigure}\hfill
\begin{subfigure}[t]{115.5pt}\centering\includegraphics[width=115.5pt]{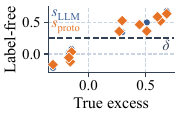}\caption{It splits the graphs at $\delta$.}\end{subfigure}
\caption{Reading polarity on eleven graphs. (a, c) LLM estimates of the homophily $H$ (\S\ref{sec:formulation}) cost 500 calls and drift from the truth. (b, d) Excess agreement along edges (\S\ref{sec:gate}) splits the graphs at $\delta$.}
\Description{Four panels in a two-by-two grid, two schematics in the top row and two scatter plots in the bottom row. First, a small graph whose sampled edges go to two LLMs that answer whether the endpoints share a class, giving one homophily estimate below a dashed true value and one above it by a drift. Second, the excess agreement, the share of edges whose endpoints carry the same label minus the chance agreement, which sets the sign to plus when it reaches delta. Third, homophily estimated by GPT-4o-mini and DeepSeek-V4-Flash against the true homophily on eleven graphs, with the two estimates on Children on either side of the diagonal and DeepSeek-V4-Flash well above it. Fourth, the excess agreement of LLM labels and of prototype predictions against the true excess agreement, with all homophilous graphs above the threshold line and all WebKB graphs below it.}
\label{fig:motiv}
\end{figure}

\begin{figure}[t]
\centering
\includegraphics[width=\linewidth]{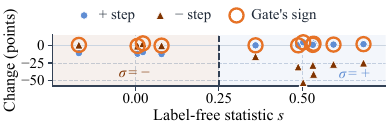}
\caption{Accuracy change from one step of Eq.~\eqref{eq:prop} on prototype scores, toward or away from the neighbours. The better sign flips where the statistic $s$ crosses $\delta$ (\S\ref{sec:gate}), and the gate picks it on every graph.}
\Description{A scatter over eleven graphs of the accuracy change from one propagation step toward the neighbours, blue dots, or away from them, brown triangles, against the label-free statistic s, with a vertical dashed line at delta. Left of the line lie the four WebKB graphs, where the step toward the neighbours loses about ten points and the step away loses little or gains. Right of it lie the seven homophilous graphs, where the step toward the neighbours gains a little and the step away loses much more. Orange rings mark the sign that the gate picks, which is the better sign on every graph.}
\label{fig:step}
\end{figure}

However, label propagation and GNNs pull a node toward its neighbours, because they assume that linked nodes tend to share a class~\cite{mcpherson}.
On hyperlink graphs this assumption fails~\cite{h2gcn,geomgcn}, and one propagation step toward the neighbours costs at least \nPlusWebMinAbs points of accuracy on each WebKB graph (Figure~\ref{fig:step}).
We call the sign that links should carry into a prediction \emph{\polarity}, and we find that it can be read without any true label.
The agreement of LLM labels along edges in excess of chance~\cite{kappa,platonov24}, and the same quantity for text-prototype predictions, exceed 0.25 on all seven homophilous graphs of our benchmark and stay below it on all four WebKB graphs (Figure~\ref{fig:motiv}d), and one propagation step toward the neighbours helps exactly on the former (\S\ref{sec:motiv}).

Two problems stand between \polarity and a zero-shot method.
\textbf{(C1)} \emph{How can a method set the sign of structure on a new graph without any true label?}
Asking an LLM for the homophily costs extra calls and inherits the LLM's bias (\S\ref{sec:motiv}).
The released code of LLM-BP (appr.), its approximate variant, reads the sign from a hand-written list of heterophilous graphs, which does not include a new graph.
Counting how often linked nodes share an LLM label mixes homophily with chance agreement, which depends on the number and sizes of the classes, and label noise lowers the count further.
\textbf{(C2)} \emph{How can noisy LLM labels, text and signed structure be combined when the number of labels varies?}
Propagating raw labels cannot repair errors that linked pages share, and on a graph whose links join different classes it pulls pages toward wrong classes.
A GNN trained on the labels needs them on most nodes and learns its aggregation from the noisy LLM labels.

\begin{figure*}[t]
\centering
\includegraphics[width=\textwidth]{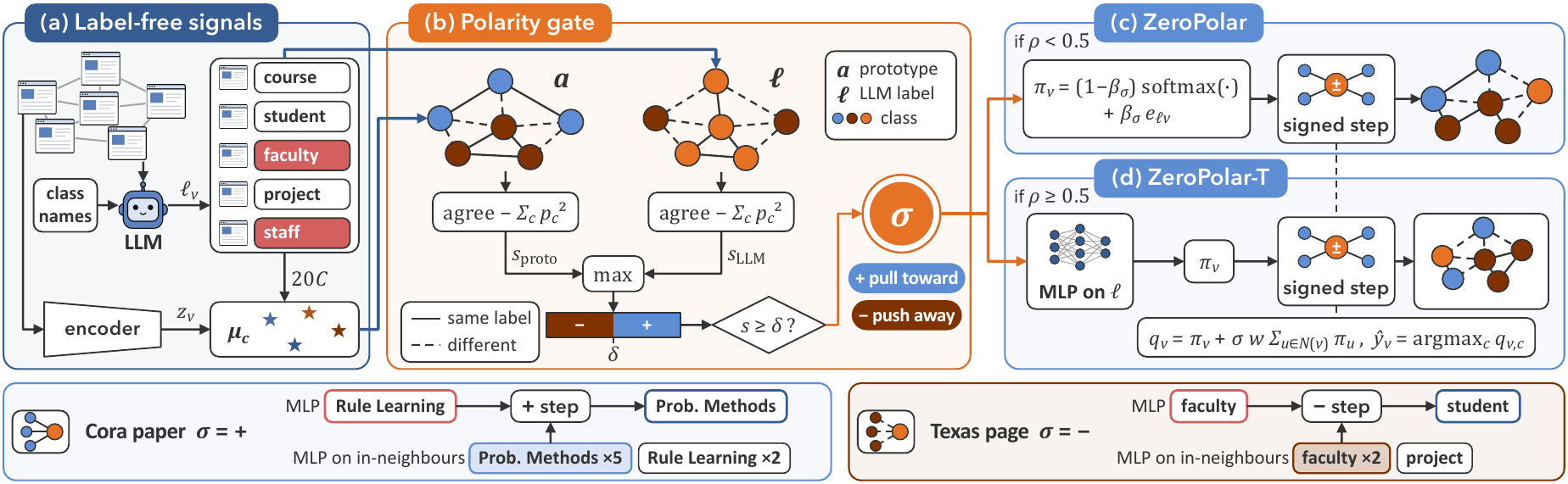}
\caption{Overview of \sys. The gate (b) reads $\sigma$ (\S\ref{sec:gate}) from the signals in (a), and the budget routes it to \sys (c) or \sysT (d). The bottom row shows two measured cases, on Cora and Texas.}
\Description{Framework diagram in four panels and a bottom row. Panel a shows a graph of web pages and an LLM that reads each page with the class names and labels it with a class, some of them wrong, and a text encoder whose embeddings meet class prototypes built from twenty LLM labels per class, giving the prototype prediction of each node. Panel b draws the prototype predictions and the LLM labels on two small graphs with solid same-label and dashed different-label edges, computes the excess agreement of each, takes their maximum and compares it with delta, which sets the sign sigma, plus to pull toward the neighbours and minus to push away. A bus carries sigma to panel c, \sys{}, if fewer than half of the test nodes carry an LLM label, and to panel d, \sysT{}, otherwise. \sys{} mixes softmax scores with the one-hot LLM label, while \sysT{} trains an MLP on the LLM labels. Both apply the same signed step, whose formula sits in panel d, and predict the class with the highest score. The bottom row shows a Cora paper whose MLP label Rule Learning becomes Probabilistic Methods after a plus step, since the MLP gives most of its in-neighbours Probabilistic Methods, and a Texas page whose MLP label faculty becomes student after a minus step that pushes it away from the faculty label of most of its in-neighbours.}
\label{fig:overview}
\end{figure*}

To this end, we propose \sys (zero-shot polarity propagation), shown in Figure~\ref{fig:overview}.
To address C1, a polarity gate computes the excess agreement of the LLM labels and of the prototype predictions along edges and sets the sign of structure to plus when the larger one reaches a threshold~$\delta$.
For C2, the sign drives one propagation step in two implementations matched to the labelling budget.
Training-free \sys mixes prototype scores with whatever LLM labels exist, and \sysT, used once at least half of the nodes carry a label, trains a light text classifier on the labels and propagates its predictions.
We validate both on eleven text-attributed graphs, two additional graphs and three further Web platforms, with five LLMs and \nEncW encoder settings.
With one GPT-4o label per node, \sysT reaches the highest eleven-graph mean (\nAllT\%), \nDiffBP points above LLM-BP and significantly above all \nNbaseW baselines (\S\ref{sec:main}).
With labels on \nBudPct\% of the nodes, \sys reaches \nBudZ\% against \nBudSage\% for a GNN trained on as many randomly chosen labels, a lead of \nBudGap points, while also using twenty prototype labels per class (\S\ref{sec:budget}).
This paper makes the following three contributions.
\begin{itemize}
\item We name \polarity on text-attributed Web graphs and show that a label-free excess agreement separates all eleven graphs by the better sign of structure, which turns the choice between trusting and distrusting links into a measurement that needs no list of graph families.
\item We propose \sys, which combines a polarity gate that sets the sign of structure from LLM labels and text prototypes without any true label (C1) with one signed propagation step in two implementations, one for few labels and one for many (C2).
\item We compare \sys with \nNbaseW baselines and six published graph models on eleven graphs, test it on two additional graphs, and study controls, label budgets, label noise, LLMs, encoders, few-shot labels, further platforms, cost and two case studies.
\end{itemize}

\section{Related Work}
\label{sec:related}

\paragraph{Zero-shot inference on text-attributed graphs.}
LLMs serve graphs as text enhancers, predictors or aligners~\cite{llmgraphsurvey,chenexplore,gfmsurvey}.
The simplest zero-shot baselines read the text alone, by asking an LLM such as GPT-4o~\cite{gpt4o} or by matching the embeddings of an encoder such as bge-en-icl~\cite{bgeicl} or LLM2Vec~\cite{llm2vec} to class embeddings.
Graph foundation models such as OFA~\cite{ofa}, ZeroG~\cite{zerog}, UniGLM~\cite{uniglm}, GOFA~\cite{gofa}, GraphCLIP~\cite{graphclip}, UniGraph~\cite{unigraph}, AnyGraph~\cite{anygraph}, UniGTE~\cite{unigte}, RAG-GFM~\cite{raggfm} and the Universal Classifier~\cite{uc} pretrain on source graphs and transfer to new ones.
GraphGPT~\cite{graphgpt}, LLaGA~\cite{llaga} and TEA-GLM~\cite{teaglm} map graphs into the token space of an LLM, and Graph-R1~\cite{graphr1} trains an LLM to reason over graphs.
Text-space benchmarks~\cite{textspace} evaluate such models across domains, and on unseen graphs these models often trail simple text baselines~\cite{llmbp}.
Other models transfer through hyperbolic retrieval~\cite{hyrag}, hyperbolic alignment~\cite{h4g} or motif contrast with subgraph alignment~\cite{graspllm}, and all of them train on source graphs.

Quantized graph tokens let LLMs read structure~\cite{qtag}, and HS2C compresses the context an LLM reads into homophilous communities~\cite{hs2c}.
ZPT~\cite{zpt} prompt-tunes a graph-language model pretrained on the target graph, and TAG~\cite{tabfm} has tabular foundation models classify nodes in context from labelled nodes of the same graph.
LLM-BP~\cite{llmbp} is training-free and propagates beliefs with a homophily that GPT-4o-mini~\cite{gpt4omini} estimates from sampled edges.
Dynamic bundling~\cite{dynbundle} queries the LLM with bundles of nodes, and GraphSearch~\cite{graphsearch} and graph walks~\cite{cny} let an LLM fetch neighbours at several calls per node.
Retrieval-based message passing~\cite{rta} recasts aggregation as retrieval.
\sys instead decides from the graph itself, at no extra LLM call, whether the neighbours of a node should be trusted.

\paragraph{LLM labels for node classification.}
LLM-GNN~\cite{llmgnn} has an LLM annotate selected nodes and trains a GNN on the labels, and Locle~\cite{locle} adds active self-training that sends hard nodes to the LLM, keeps GNN pseudo-labels for stable ones and corrects labels on a rewired graph.
We compare with both.
CANE~\cite{cane} estimates how reliable LLM labels are in each feature cluster to decide which pseudo-labels to trust, whereas \sys decides how the links enter.
GLANCE~\cite{glance} learns from true labels when to ask an LLM to refine a GNN.
Co-teaching~\cite{coteach} lets an LLM and a GNN teach each other from a few true labels per class, whereas \sys uses no true label.
Studies ask when LLMs help node classification~\cite{llmnodebed,chenexplore} and how they cope with heterophilous graphs~\cite{llm4heg}, while TAPE~\cite{tape}, GLEM~\cite{glem} and GraphAdapter~\cite{graphadapter} combine language models with GNNs using true labels.
These methods decide which labels to trust or train message passing that assumes agreeing neighbours.
Our baselines adapt LLM-GNN to a label on every node and train a multilayer perceptron (MLP), GCN and GraphSAGE on the labels, whereas \sysT trains only a text classifier and leaves the sign of structure to the gate.

\paragraph{Homophily and signed propagation.}
Homophily, the tendency of linked nodes to share a class~\cite{mcpherson}, underlies label propagation~\cite{lp} and GNNs~\cite{gcn,graphsage}, and the WebKB graphs~\cite{webkb,geomgcn} show where it fails.
Heterophily GNNs separate ego and neighbour embeddings~\cite{h2gcn}, learn signed propagation weights~\cite{gprgnn,fagcn,ggcn}, mix filter channels~\cite{acmgcn} or polarize messages~\cite{polarized}, and evaluation studies revisit their benchmarks and measures~\cite{platonov23,platonov24}.
Adjusted homophily~\cite{platonov24} removes the agreement expected by chance from edge agreement, in the same way as Cohen's kappa for two raters~\cite{kappa}.
These GNNs train on labels, here from the LLM, and the measures need true labels.
\sys computes this excess agreement from LLM labels and prototype predictions instead, and uses it only to set the sign of one step.

\section{Preliminaries and Structural Polarity}
\label{sec:prelim}

\subsection{Preliminaries}
\label{sec:formulation}

\paragraph{Zero-shot node classification.}
A text-attributed graph $G=(V,E)$ has a text $x_v$ on every node and edges $E$, and $\mathcal{N}(v)$ denotes the in-neighbours of node $v$.
Each node belongs to one of $C$ classes whose names are known, and no node has a true label.
A method predicts a class $\hat y_v$ for each of $N$ test nodes from the texts, the edges, the class names and calls to an LLM, and it is evaluated by accuracy on these nodes.

\paragraph{LLM labels and prototypes.}
An LLM reads $x_v$ and the class names and returns a label $\ell_v$ at the cost of one call.
A budget buys labels for a set $L$ of $\rho N$ test nodes, and the labels are noisy, since GPT-4o labels taken as predictions reach \nAllLLM\% on our eleven graphs (\S\ref{sec:main}).
As in LLM-BP~\cite{llmbp}, a frozen encoder $f$ maps each text to $z_v=f(x_v)$ and the LLM labels a sample of $20C$ nodes, whose embeddings, averaged per label, give our class prototypes $\mu_c$.

\paragraph{Homophily and propagation.}
The edge homophily $H$ is the share of edges whose endpoints share a class~\cite{h2gcn}.
Given class scores $\pi_v$ in the probability simplex, one propagation step computes:
\begin{equation}\label{eq:prop}
q_v=\pi_v+w\sum_{u\in\mathcal{N}(v)}\pi_u ,
\end{equation}
and predicts $\arg\max_c q_{v,c}$.
The step moves $v$ toward the classes of its neighbours, and on prototype scores it is the step of LLM-BP (appr.)~\cite{llmbp}.
Label propagation~\cite{lp} and the layers of a GNN~\cite{gcn,graphsage} exert the same pull on every graph, and the pull helps only on graphs whose $H$ is high.

\subsection{Structural Polarity}
\label{sec:motiv}

\paragraph{The sign of structure decides whether it helps.}
We score each node by its prototypes and apply one step of Eq.~\eqref{eq:prop} toward its neighbours or, with $-w$, away from them.
On the seven homophilous graphs the step toward the neighbours helps and the step away costs at least \nMinusHomoMinAbs points, while on the four WebKB graphs the step toward the neighbours costs at least \nPlusWebMinAbs points (Figure~\ref{fig:step}).
We call the better of the two signs on a graph its \emph{\polarity}.
\emph{A zero-shot method therefore has to choose the sign before it uses the links}, which motivates the polarity gate (\S\ref{sec:gate}).

\paragraph{LLM estimates of homophily drift with the annotator.}
LLM-BP sets the strength of its propagation from an $H$ that GPT-4o-mini~\cite{gpt4omini} estimates from 500 LLM calls on sampled edges (Figure~\ref{fig:motiv}a).
The estimate moves with the LLM, and DeepSeek-V4-Flash~\cite{deepseekv4}, given the same prompts, puts Children at \nDsHChild against a true \nTrueHChild (Figure~\ref{fig:motiv}c).
\emph{A sign read from LLM judgments of homophily inherits this drift and costs extra calls}, so the gate reads only labels a method already holds (\S\ref{sec:gate}).

\paragraph{Excess agreement separates the graphs without labels.}
The share of edges whose endpoints carry the same LLM label mixes homophily with chance agreement, which grows when one class dominates.
We subtract the chance agreement $\sum_c p_c^2$ of the label shares $p_c$, in the spirit of Cohen's kappa~\cite{kappa} and adjusted homophily~\cite{platonov24} (Figure~\ref{fig:motiv}b).
The larger of this excess agreement on GPT-4o labels and on prototype predictions is at least \nSHomoMin on every homophilous graph and at most \nSWebMax on every WebKB graph, and both track the excess agreement of the true labels (Figure~\ref{fig:motiv}d).
\emph{Polarity can thus be read without any true label}, and the polarity gate reads both (\S\ref{sec:gate}).

\section{ZeroPolar}
\label{sec:method}

\subsection{Overview}
\label{sec:overview}
Figure~\ref{fig:overview} summarizes how \sys turns \polarity into predictions.
In (a), an LLM labels nodes from their text and a frozen encoder scores every node against the class prototypes.
In (b), the polarity gate measures the excess agreement of both along edges and sets the sign $\sigma\in\{+1,-1\}$ of structure, which answers C1.
A budget switch then routes $\sigma$ to \sysT in (d) when at least half of the test nodes carry an LLM label, $\rho\ge0.5$, and to training-free \sys in (c) otherwise.
Both implementations end with the same signed step, so the switch and the step answer C2.

\subsection{Polarity Gate}
\label{sec:gate}
Let $E_L$ be the edges whose endpoints both carry an LLM label and $p_c$ the share of LLM labels equal to $c$.
The gate computes the excess agreement of the LLM labels:
\begin{equation}\label{eq:sllm}
s_{\mathrm{LLM}}=\frac{1}{|E_L|}\sum_{(u,v)\in E_L}\mathbf{1}[\ell_u=\ell_v]-\sum_{c=1}^{C}p_c^2 ,
\end{equation}
and the same quantity $s_{\mathrm{proto}}$ for the prototype predictions $a_v=\arg\max_c\cos(z_v,\mu_c)$ over all edges, with $p_c$ the share of predictions equal to $c$.
It then sets the sign of structure:
\begin{equation}\label{eq:gate}
\sigma=\begin{cases}+1 & \text{if } s=\max(s_{\mathrm{LLM}},s_{\mathrm{proto}})\ge\delta,\\ -1 & \text{otherwise,}\end{cases}
\end{equation}
with $\delta=0.25$ on every graph.

\paragraph{Chance correction removes the agreement that class sizes alone produce.}
Random edges have zero excess agreement whatever the number and sizes of the classes, so a dominant class does not inflate $s_{\mathrm{LLM}}$ or $s_{\mathrm{proto}}$.
Label noise that hits nodes independently and uniformly scales the excess agreement toward zero in expectation without changing its sign.

\begin{proposition}\label{prop:noise}
Let each node of clean class $y$ keep its class with probability $1-\eta$ and otherwise take a class drawn uniformly from the $C$ classes, independently across nodes.
If $\gamma$ is the excess agreement of the clean labels, the noisy labels have excess agreement $(1-\eta)^2\gamma$ in expectation, with the chance term taken at the expected class shares.
\end{proposition}

\begin{proof}
Write $J_c$ for the share of edges with both endpoints in class $c$ and $p_c$ for the share of nodes in class $c$, as in Eq.~\eqref{eq:sllm}, so $\gamma=\sum_c J_c-\sum_c p_c^2$.
A noisy label equals $c$ with probability $(1-\eta)\mathbf{1}[y=c]+\eta/C$, independently at the two endpoints of an edge.
Summing over the classes, the edge agreement $\tilde A$ and the chance agreement $\tilde B$ of the noisy labels are:
\begin{equation}
\tilde A=(1-\eta)^2\textstyle\sum_c J_c+\epsilon,\qquad \tilde B=(1-\eta)^2\textstyle\sum_c p_c^2+\epsilon,
\end{equation}
where $\epsilon=2(1-\eta)\eta/C+\eta^2/C$ collects the terms that noise adds to both, and $\tilde A-\tilde B=(1-\eta)^2\gamma$.
\end{proof}

In expectation, such noise can thus push a homophilous graph below $\delta$ but cannot lift a heterophilous graph above it.
Taking the larger of two statistics with different sources of error guards against this one-sided error, since the prototype predictions rest only on the $20C$ prototype labels and not on the labels of the test nodes.

\subsection{One Signed Step in Two Implementations}
\label{sec:step}
Both implementations answer C2 by building class scores $\pi_v$ for every node and then applying one step of Eq.~\eqref{eq:prop} with the sign that the gate sets:
\begin{equation}\label{eq:step}
q_v=\pi_v+\sigma w\sum_{u\in\mathcal{N}(v)}\pi_u,\qquad \hat y_v=\arg\max_c q_{v,c},
\end{equation}
with $w=0.5$.
With $\sigma=+1$ a node moves toward the classes of its neighbours, and with $\sigma=-1$ it moves away from them, so a page whose neighbours all look like faculty pages is pushed away from the faculty class toward the classes its text supports.

\paragraph{\sys for few labels.}
Training-free \sys mixes the prototype scores with the LLM labels:
\begin{equation}\label{eq:prior}
\pi_v=(1-\beta_\sigma)\,\mathrm{softmax}_c\big(\cos(z_v,\mu_c)/\tau\big)+\beta_\sigma\,\mathbf{e}_{\ell_v},
\end{equation}
where $\mathbf{e}_{\ell_v}$ is the one-hot vector of the label of $v$, zero if $v$ has none, and $\tau=0.01$.
The label weight is $\beta_{+}=0.8$ on graphs gated $+1$ and $\beta_{-}=0.3$ on graphs gated $-1$, both chosen on the five design graphs listed in \S\ref{sec:setup}.

\paragraph{\sysT for many labels.}
When at least half of the test nodes carry an LLM label, \sysT trains a two-layer MLP $g$ on the labelled pairs $(z_v,\ell_v)$ and sets $\pi_v=\mathrm{softmax}(g(z_v))$ before the step of Eq.~\eqref{eq:step}.
The MLP sees no edges, so it learns how texts map to labels and overrides labels that disagree with similar texts, while the gate alone decides how the links enter.
It stops early on a held-out tenth of the LLM labels.

\paragraph{One signed step keeps the text in charge.}
One step adds the scores of each neighbour once, so the text of a node still decides when its neighbours disagree, and a second step lowers the eleven-graph mean accuracy of both implementations (\S\ref{sec:abl}).

\paragraph{Cost.}
Both implementations make $20C+\rho N$ LLM calls, $20C$ of them for the prototypes as in LLM-BP.
The gate and the step take $O(|E|C)$ time, and \sysT adds the training of one small MLP, which takes seconds on a GPU (\S\ref{sec:cost}).

\section{Experiments}
\label{sec:eval}

\begin{table*}[t]
\centering
\small
\setlength{\tabcolsep}{2pt}
\caption{Zero-shot accuracy (\%) with GPT-4o labels on eleven graphs.}
\label{tab:main}
\begin{tabular}{@{}c@{\hspace{3pt}}l c ccccc cccccc ccc@{}}
\toprule
 & & & \multicolumn{5}{c}{Design graphs} & \multicolumn{6}{c}{Other graphs} & \multicolumn{3}{c}{Mean}\\
\cmidrule(lr){4-8}\cmidrule(lr){9-14}\cmidrule(l){15-17}
 & Method & Calls & Cora & Cite & Pub & Wiki & Corn & Hist & Child & Sport & Tex & Wisc & Wash & Design & Other & All\\
\midrule
 & Vanilla~\cite{bgeicl} & 0 & 68.610 & 67.099 & 76.363 & 63.171 & 85.288 & 58.335 & 28.940 & 59.372 & 78.342 & 75.572 & 66.783 & 72.106 & 61.224 & 66.171\\
 & LLM direct~\cite{gpt4o} & $N$ & 70.295 & 64.770 & 89.858 & 66.102 & 45.550 & 53.303 & 30.764 & 66.358 & 63.102 & 56.604 & 48.908 & 67.315 & 53.173 & 59.601\\
\midrule
\multirow{6}{*}[-2.1pt]{\rotatebox[origin=c]{90}{Training-free}} & LLM labels + LP~\cite{lp} & $N$ & \underline{75.092} & 68.355 & \textbf{91.202} & \underline{72.738} & 48.168 & \underline{61.858} & \underline{32.800} & \underline{71.596} & 67.914 & 58.491 & 51.092 & 71.111 & 57.292 & 63.573\\
 & LLM-BP~\cite{llmbp} & $20C{+}500$ & 72.589 & \underline{69.508} & 75.547 & 67.746 & 83.264 & 59.860 & 24.807 & 61.920 & \underline{81.658} & \textbf{77.748} & \textbf{73.144} & 73.731 & 63.190 & 67.981\\
 & LLM-BP (appr.)~\cite{llmbp} & $20C$ & 71.415 & 68.658 & 76.813 & 67.964 & 84.921 & 59.490 & 29.402 & 61.512 & 79.394 & \underline{75.648} & 70.044 & 73.954 & 62.582 & 67.751\\
 & Gated switch & $20C{+}N$ & \underline{75.092} & 68.355 & \textbf{91.202} & \underline{72.738} & \textbf{85.288} & \underline{61.858} & \underline{32.800} & \underline{71.596} & 78.342 & 75.572 & 66.783 & \underline{78.535} & \underline{64.492} & \underline{70.875}\\
 & GOFA-T~\cite{gofa} & 0 & 61.993 & 65.650 & 68.250 & 67.650 & 58.115 & 26.350 & 17.350 & 37.050 & 52.406 & 50.943 & 31.004 & 64.332 & 35.851 & 48.797\\
\cmidrule(l){2-17}
 & \sys & $20C{+}N$ & \textbf{76.052} & \textbf{69.855} & \underline{91.136} & \textbf{73.435} & \underline{85.183} & \textbf{63.574} & \textbf{33.151} & \textbf{73.343} & \textbf{82.478} & 75.585 & \underline{71.907} & \textbf{79.132} & \textbf{66.673} & \textbf{72.336}\\
\midrule
\multirow{7}{*}[-2.1pt]{\rotatebox[origin=c]{90}{Trained}} & SAGE h64~\cite{graphsage} & $N$ & 77.798 & 70.577 & 89.571 & 74.922 & 73.647 & 67.986 & \textbf{35.660} & \underline{77.704} & 82.888 & 73.208 & 67.249 & 77.303 & 67.449 & 71.928\\
 & GPR-GNN h256~\cite{gprgnn} & $N$ & \underline{78.290} & 70.551 & 88.151 & \underline{75.891} & 60.558 & 66.855 & 35.428 & \textbf{78.351} & 72.371 & 66.038 & 64.629 & 74.688 & 63.945 & 68.828\\
 & H2GCN h64~\cite{h2gcn} & $N$ & 76.937 & \underline{70.668} & \underline{90.357} & 74.796 & 72.077 & 68.520 & \underline{35.647} & 77.290 & 78.966 & 72.704 & \underline{73.799} & 76.967 & 67.821 & 71.978\\
 & FAGCN h256~\cite{fagcn} & $N$ & 77.122 & 70.642 & 88.937 & 74.523 & 71.030 & \underline{69.394} & 34.942 & 77.323 & 73.797 & \textbf{75.597} & 72.198 & 76.450 & 67.208 & 71.409\\
 & ACM-GCN h64~\cite{acmgcn} & $N$ & 77.860 & 70.369 & \textbf{91.176} & 74.158 & \underline{76.789} & 67.946 & 35.313 & 77.217 & \underline{87.522} & 72.579 & 72.926 & \underline{78.070} & \underline{68.917} & \underline{73.078}\\
 & Locle (zero-shot cache)~\cite{locle} & $50C$ & \textbf{78.352} & 68.342 & 82.590 & 69.534 & 44.852 & 65.483 & 27.456 & 48.795 & 69.697 & 56.855 & 42.649 & 68.734 & 51.823 & 59.510\\
\cmidrule(l){2-17}
 & \sysT & $20C{+}N$ & 77.921 & \textbf{71.070} & 90.112 & \textbf{75.953} & \textbf{78.360} & \textbf{69.775} & 35.040 & 77.336 & \textbf{88.235} & \underline{74.717} & \textbf{75.400} & \textbf{78.683} & \textbf{70.084} & \textbf{73.993}\\
\bottomrule
\end{tabular}
\end{table*}

\subsection{Setup}
\label{sec:setup}

\paragraph{Graphs and annotators.}
We use the eleven text-attributed graphs of LLM-BP~\cite{llmbp}, namely Cora, Citeseer and Pubmed~\cite{sen2008,planetoid}, WikiCS~\cite{wikics}, History, Children and Sportsfit~\cite{cstag}, and the WebKB hyperlink graphs Cornell, Texas, Wisconsin and Washington~\cite{webkb,geomgcn}, with the test nodes of its splits, so $N$ is the number of labelling calls per graph.
Five design graphs, Cora, Citeseer, Pubmed, WikiCS and Cornell, guided the development of both implementations, and the other six are reported separately.
Computer~\cite{cstag} and a 30k-node subgraph of arXiv~\cite{ogb} are two additional graphs beyond the eleven, and Instagram, Reddit~\cite{graphadapter} and Photo~\cite{cstag} add three further Web platforms with social and product content.
GPT-4o~\cite{gpt4o} labels the eleven graphs with the prompt and parser of LLM-BP, and Qwen3-8B~\cite{qwen3}, Llama-3.1-8B~\cite{llama3}, Mistral-7B~\cite{mistral} and Qwen3.8-27B~\cite{qwen38} serve as open annotators.
The encoder is bge-en-icl~\cite{bgeicl} with the task-aware prompt of LLM-BP.
App.~\ref{app:setup} gives the parsers, label budgets, noise and training recipes.

\paragraph{Baselines.}
We run \nNbaseW{} baselines under one protocol.
Without the graph, Vanilla classifies each node by its text embedding alone, and LLM direct takes the LLM label.
With a fixed use of structure, LLM labels + LP propagates the LLM labels~\cite{lp}, LLM-BP~\cite{llmbp} runs belief propagation with a homophily that GPT-4o-mini~\cite{gpt4omini} estimates from 500 calls, and LLM-BP (appr.) applies the step of Eq.~\eqref{eq:prop} to prototype scores with the sign that its released code fixes per graph family.
A gated switch applies LLM labels + LP on graphs that our gate sets to $+1$ and Vanilla otherwise.
Adapting LLM-GNN~\cite{llmgnn} to label every test node, we train MLP, GCN~\cite{gcn} and GraphSAGE~\cite{graphsage} with 64 or 256 hidden units on the LLM labels, and Table~\ref{tab:main} shows SAGE h64, the variant chosen on the design graphs.
Four heterophily GNNs, GPR-GNN~\cite{gprgnn}, H2GCN~\cite{h2gcn}, FAGCN~\cite{fagcn} and ACM-GCN~\cite{acmgcn}, train on the same labels, and Locle~\cite{locle} spends $B=50C$ LLM calls on test nodes chosen by its active selection (App.~\ref{app:locle}).
GOFA-T~\cite{gofa}, the instruction-tuned version of the graph foundation model GOFA, answers zero-shot with its released weights and reads no LLM labels.

\paragraph{Protocol.}
Training-free rows average over 30 prototype seeds of LLM-BP and trained rows over three training seeds.
Differences are paired over nodes with bootstrap~\cite{efron79} 95\% confidence intervals, and a difference is significant when its interval excludes zero.
In the tables, Other is the other six graphs and Addl.\ the two additional graphs, gray rows are controls, and bold and underline mark the top two of each block.

\subsection{Main Results}
\label{sec:main}

\begin{table}[t]
\centering
\small
\setlength{\tabcolsep}{3pt}
\caption{Accuracy (\%) with Qwen3.8-27B labels on two additional graphs, Computer and arXiv.}
\label{tab:fresh}
\begin{tabular*}{\linewidth}{@{}c@{\hspace{3pt}}l@{\extracolsep{\fill}\hspace{\tabcolsep}}ccc@{}}
\toprule
 & Method & Computer & arXiv & Mean\\
\midrule
 & Vanilla~\cite{bgeicl} & 36.41 & 67.43 & 51.92\\
 & LLM direct & 62.10 & 75.46 & 68.78\\
\midrule
\multirow{6}{*}[-2.1pt]{\rotatebox[origin=c]{90}{Training-free}} & LLM labels + LP~\cite{lp} & \underline{63.92} & \underline{75.46} & \underline{69.69}\\
 & LLM-BP~\cite{llmbp} & 51.99 & 66.37 & 59.18\\
 & LLM-BP (appr.)~\cite{llmbp} & 46.85 & 67.70 & 57.27\\
 & Gated switch & \underline{63.92} & \underline{75.46} & \underline{69.69}\\
 & GOFA-T~\cite{gofa} & 57.45 & 57.40 & 57.42\\
\cmidrule(l){2-5}
 & \sys & \textbf{64.69} & \textbf{75.50} & \textbf{70.10}\\
\midrule
\multirow{8}{*}[-2.1pt]{\rotatebox[origin=c]{90}{Trained}} & MLP h256~\cite{llmgnn} & 65.59 & 77.19 & 71.39\\
 & SAGE h256~\cite{graphsage} & 70.58 & \textbf{78.31} & 74.45\\
 & SAGE h64~\cite{graphsage} & 71.15 & 76.67 & 73.91\\
 & GPR-GNN h256~\cite{gprgnn} & \underline{75.37} & 76.23 & \underline{75.80}\\
 & H2GCN h64~\cite{h2gcn} & 70.77 & \underline{77.55} & 74.16\\
 & FAGCN h256~\cite{fagcn} & 72.18 & 76.69 & 74.44\\
 & ACM-GCN h64~\cite{acmgcn} & 65.71 & 77.21 & 71.46\\
\cmidrule(l){2-5}
 & \sysT & \textbf{77.08} & 77.01 & \textbf{77.05}\\
\bottomrule
\end{tabular*}
\end{table}

\begin{figure}[t]
\centering
\begin{subfigure}[t]{0.509\linewidth}\centering\includegraphics[width=\linewidth]{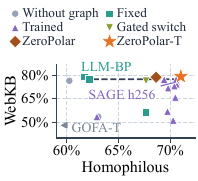}\caption{Two graph families.}\end{subfigure}\hfill
\begin{subfigure}[t]{0.480\linewidth}\centering\includegraphics[width=\linewidth]{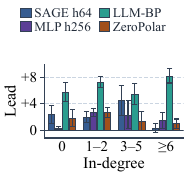}\caption{Lead by in-degree.}\end{subfigure}
\caption{Mean accuracy on the homophilous against the WebKB graphs, where the dashed line joins the baselines that no other beats on both (a), and the lead of \mbox{\sysT} by in-degree with 95\% CIs (b), with GPT-4o labels.}
\Description{Two panels. Left, a scatter of the homophilous-graph mean against the WebKB mean, with baselines marked by family and a dashed line through LLM-BP, LLM-BP (appr.), ACM-GCN h64 and SAGE h256, which no other baseline beats on both axes. \sysT{}, a star near 71 and 79 percent, lies above and to the right of every baseline, \sys{} is close to it, and GOFA-T lies left of the frame. Right, bars for the lead of \sysT{} in four in-degree buckets over SAGE h64, MLP h256, LLM-BP and \sys{} with interval bars, about 5 to 8 points over LLM-BP in every bucket and near zero over SAGE h64 at six or more in-edges.}
\label{fig:lead}
\end{figure}

\begin{table}[t]
\centering
\small
\setlength{\tabcolsep}{3pt}
\caption{Controls that remove or flip one part of each implementation, GPT-4o labels.}
\label{tab:abl}
\renewcommand{\arraystretch}{1.08}
\begin{tabular*}{\linewidth}{@{}c@{\hspace{3pt}}l@{\extracolsep{\fill}\hspace{\tabcolsep}}ccc@{}}
\toprule
 & Variant & Design & Other & All\\
\midrule
\multirow{4}{*}[-2.1pt]{\rotatebox[origin=c]{90}{\sysT}} & No step (MLP only) & 76.69 & 69.44 & 72.74\\
 & Step always $+$ & 74.39 & 64.11 & 68.78\\
 & Step always $-$ & 42.47 & 55.28 & 49.45\\
\cmidrule(l){2-5}
 & \textbf{\sysT} & 78.68 & 70.08 & 73.99\\
\midrule
\multirow{7}{*}[-2.1pt]{\rotatebox[origin=c]{90}{\sys}} & No gate (always $+$) & 74.47 & 60.18 & 66.68\\
 & No step & 77.95 & 65.20 & 71.00\\
 & No LLM labels & 73.95 & 62.58 & 67.75\\
 & LLM labels only & 75.48 & 56.38 & 65.06\\
 & Gate reads $s_{\mathrm{LLM}}$ only & 79.13 & 66.67 & 72.34\\
 & Gate reads $s_{\mathrm{proto}}$ only & 79.13 & 66.67 & 72.34\\
\cmidrule(l){2-5}
 & \textbf{\sys} & 79.13 & 66.67 & 72.34\\
\bottomrule
\end{tabular*}
\end{table}

\begin{table}[t]
\centering
\small
\setlength{\tabcolsep}{4pt}
\caption{Macro-F1 averaged over seeds and then over graphs.}
\label{tab:f1}
\begin{tabular*}{\linewidth}{@{\extracolsep{\fill}}lccccc@{}}
\toprule
 & \multicolumn{2}{c}{GPT-4o} & \multicolumn{3}{c}{Qwen3-8B}\\
\cmidrule(lr){2-3}\cmidrule(l){4-6}
Method & All & Other & All & Other & Addl.\\
\midrule
LLM direct & 48.98 & 43.79 & 50.10 & 43.19 & 27.41\\
\midrule
LLM-BP~\cite{llmbp} & 58.54 & 49.27 & 49.15 & 42.63 & 26.89\\
Gated switch & \underline{62.75} & \underline{53.09} & \underline{52.75} & \underline{42.73} & \textbf{30.52}\\
\cmidrule(l){1-6}
\sys & \textbf{63.38} & \textbf{53.56} & \textbf{55.52} & \textbf{46.27} & \underline{29.65}\\
\midrule
SAGE h64~\cite{graphsage} & \underline{62.71} & \underline{53.74} & \underline{54.75} & \underline{45.10} & \underline{30.35}\\
\cmidrule(l){1-6}
\sysT & \textbf{65.01} & \textbf{56.65} & \textbf{57.47} & \textbf{47.79} & \textbf{33.56}\\
\bottomrule
\end{tabular*}
\end{table}

\paragraph{With one GPT-4o label per node, \sysT has the highest eleven-graph mean.}
It reaches \nAllT\% and is significantly above each of the \nNbaseW baselines, by \nDiffBP points over LLM-BP and \nDiffSage points over SAGE h64, the LLM-GNN variant chosen on the design graphs (Table~\ref{tab:main}, and Table~\ref{tab:trainedfull} for all six variants).
Its mean is also far above the six published zero-shot graph models whose accuracy LLM-BP reports, the best of which, GOFA, reaches \nPubBest\% (Table~\ref{tab:pub}).
Over ten training seeds it leads SAGE h64 by \nSeedAll points (Table~\ref{tab:lead}).
\sysT also has the best Macro-F1 in every column of Table~\ref{tab:f1}, with \nFoneT\% on the eleven graphs with GPT-4o labels.

\paragraph{Among methods that train nothing, \sys has the best eleven-graph mean.}
\sys reaches \nAllZ\% against \nAllBP\% for LLM-BP, is significantly above GOFA-T and all six baselines that train nothing, and has the best mean on the design graphs (Table~\ref{tab:main}).

\paragraph{\sysT leads on both graph kinds.}
Its mean is \nRegHomT\% on the homophilous graphs and \nRegWebT\% on the WebKB graphs, above every baseline on both (Figure~\ref{fig:lead}a).
No baseline leads on both, since SAGE h256 is the best of them on the homophilous graphs and LLM-BP on the WebKB graphs.

\paragraph{On two additional graphs with Qwen3.8-27B labels, \sysT has the best mean.}
It is \nFreshDiff points above GPR-GNN h256, the strongest baseline there, \nGofaFresh points above GOFA-T on GOFA-T's evaluated nodes, and significantly above all \nNbaseFreshW baselines (Table~\ref{tab:fresh}).

\subsection{Label Budget}
\label{sec:budget}

\begin{figure}[t]
\centering
\begin{subfigure}[t]{0.495\linewidth}\centering\includegraphics[width=\linewidth]{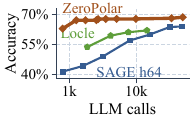}\caption{Few labels.}\end{subfigure}\hfill
\begin{subfigure}[t]{0.495\linewidth}\centering\includegraphics[width=\linewidth]{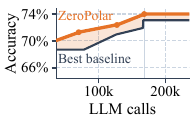}\caption{Budget switch.}\end{subfigure}
\caption{Eleven-graph mean accuracy against LLM calls, for \sys against SAGE h64 and Locle at few labels (a) and for the budget switch of \S\ref{sec:overview} against the best baseline (b), shaded where ours leads.}
\Description{Two line panels on logarithmic axes of LLM calls over the eleven graphs. Left, at few labels, \sys{} starts near 63 percent at about 800 calls with 5 prototype labels per class, reaches about 67 percent at about 1.3 thousand calls with 10, and stays near 67 to 69 percent up to about 49 thousand calls, while SAGE h64 rises from about 42 percent at about 800 calls to about 64 percent and Locle with the zero-shot cache rises from about 54 to about 62 percent over its four budgets between about 1.8 thousand and 14 thousand calls, both below \sys{} throughout. Right, from about 64 thousand to about 257 thousand calls around the full budget of one LLM label per test node, our line, which follows the budget switch between the two implementations, rises from about 70 percent to about 74 percent at the full budget and stays flat beyond it. The best baseline, the per-budget maximum of all \nNbaseW baselines, stays near 69 percent up to about 90 thousand calls and rises to about 73 percent at the full budget. Our line is above it across the whole axis, and the gap is shaded.}
\label{fig:budget}
\end{figure}

\paragraph{With few LLM labels, \sys is far ahead of a GNN trained on as many randomly chosen labels.}
With labels on \nBudPct\% of the test nodes, \sys reaches \nBudZ\% against \nBudSage\% for SAGE h64, a lead of \nBudGap points, while \sys also uses $20C$ prototype labels (Figure~\ref{fig:budget}a).
Locle (zero-shot cache), which chooses the nodes it labels, stays below \sys at all four of its budgets, even where it makes more calls.
At the smallest budgets \sys may also use five or ten prototype labels per class, the subsets that LLM-BP uses, and Figure~\ref{fig:budget}a draws the best of five, ten and twenty at each budget.
With only \nFewK prototype labels per class, \sys reaches \nFewKZ\% at about the smallest budget of SAGE h64, where SAGE h64 reaches \nFewKSage\% at this budget.

\paragraph{From half of the nodes on, \sysT is above every baseline at equal calls.}
With labels on half of the test nodes it reaches \nCrossT\% while the best baseline at equal calls reaches \nEnvCross\%, and its lead lasts up to the full budget (Figure~\ref{fig:budget}b).

\begin{table*}[t]
\centering
\small
\caption{Mean accuracy (\%) with open LLM annotators. A dash marks a setting not run.}
\label{tab:llm}
\begin{tabular*}{\textwidth}{@{}c@{\hspace{3pt}}l@{\extracolsep{\fill}\hspace{\tabcolsep}} ccccccccccc@{}}
\toprule
 & & \multicolumn{3}{c}{Qwen3.8-27B} & \multicolumn{3}{c}{Qwen3-8B} & \multicolumn{3}{c}{Llama-3.1-8B} & \multicolumn{2}{c}{Mistral-7B}\\
\cmidrule(lr){3-5}\cmidrule(lr){6-8}\cmidrule(lr){9-11}\cmidrule(l){12-13}
 & Method & Other & All & Addl. & Other & All & Addl. & Other & All & Addl. & Other & All\\
\midrule
 & Vanilla & 61.87 & 66.31 & 51.92 & 50.79 & 58.40 & 38.55 & 53.69 & 60.01 & 38.16 & 52.28 & 58.49\\
 & LLM direct & 68.91 & 72.15 & 68.78 & 53.70 & 59.34 & 44.37 & 54.12 & 58.73 & 47.27 & 53.66 & 56.69\\
\midrule
\multirow{5}{*}[-2.1pt]{\rotatebox[origin=c]{90}{Training-free}} & LLM labels + LP & \textbf{68.91} & \underline{72.15} & \underline{69.69} & \underline{53.65} & 59.33 & 44.64 & 54.07 & 58.74 & \underline{47.31} & 53.92 & 57.04\\
 & LLM-BP & 63.66 & 66.36 & 59.18 & 51.85 & 55.60 & 43.22 & \underline{57.87} & 59.31 & 41.20 & 49.10 & 45.23\\
 & LLM-BP (appr.) & 62.75 & 67.54 & 57.27 & 52.15 & \underline{60.08} & 43.98 & 56.19 & \underline{62.33} & 42.88 & 54.05 & \underline{60.47}\\
 & Gated switch & 65.43 & 69.93 & \underline{69.69} & 51.58 & 59.42 & \underline{44.89} & 56.10 & 60.85 & \underline{47.31} & \underline{54.30} & 58.59\\
\cmidrule(l){2-13}
 & \sys & \underline{68.22} & \textbf{72.27} & \textbf{70.10} & \textbf{55.96} & \textbf{62.72} & \textbf{46.68} & \textbf{59.15} & \textbf{63.52} & \textbf{48.44} & \textbf{56.69} & \textbf{61.07}\\
\midrule
\multirow{3}{*}[-2.1pt]{\rotatebox[origin=c]{90}{Trained}} & SAGE h64 & \underline{68.84} & \underline{73.07} & \underline{73.91} & \underline{57.86} & \underline{63.44} & \underline{54.54} & \underline{54.26} & \underline{59.23} & \underline{51.01} & \underline{57.03} & \underline{60.63}\\
 & Locle & -- & -- & -- & 44.97 & 53.13 & 43.73 & 40.17 & 48.78 & 49.31 & 37.69 & 46.93\\
\cmidrule(l){2-13}
 & \sysT & \textbf{69.76} & \textbf{74.22} & \textbf{77.05} & \textbf{59.27} & \textbf{65.59} & \textbf{56.73} & \textbf{56.46} & \textbf{61.37} & \textbf{53.14} & \textbf{58.58} & \textbf{62.56}\\
\bottomrule
\end{tabular*}
\end{table*}

\subsection{Controls}
\label{sec:abl}

\begin{figure}[t]
\centering
\includegraphics[width=\linewidth]{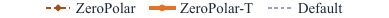}\\
\includegraphics[width=0.409\linewidth]{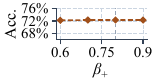}\hfill
\includegraphics[width=0.285\linewidth]{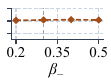}\hfill
\includegraphics[width=0.295\linewidth]{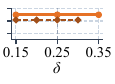}\\[1pt]
\includegraphics[width=0.409\linewidth]{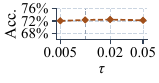}\hfill
\includegraphics[width=0.285\linewidth]{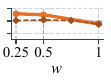}\hfill
\includegraphics[width=0.295\linewidth]{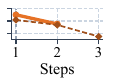}
\caption{Eleven-graph mean accuracy with GPT-4o labels when one setting of \sys or \mbox{\sysT} changes.}
\Description{Six small line panels, one per setting, beta plus, beta minus, delta, tau, w and the number of steps, on a shared accuracy axis from 66 to 76 percent. ZeroPolar stays near 72 percent across beta plus, beta minus, delta and tau and drops with more steps. ZeroPolar-T, drawn for w, delta and steps, stays near 74 percent and drops with two steps.}
\label{fig:sens}
\end{figure}

\paragraph{Without the gate, structure costs accuracy.}
Always stepping toward the neighbours costs \sysT \nCtrlPlusT points and \sys \nCtrlNoGateZ points on the eleven-graph mean, and always stepping away costs \sysT \nCtrlMinusT points (Table~\ref{tab:abl}).
The step and the LLM labels each add accuracy, since the signed step adds \nCtrlNoStepT points to the MLP of \sysT, and \sys loses \nCtrlNoPropZ points without the step and \nCtrlNoLabelZ points without the LLM labels, while keeping only the labels loses more.
Without the LLM labels, \sys matches LLM-BP (appr.) on all eleven graphs, because its gate finds, without any list of graph families, the sign that the released code of LLM-BP (appr.) fixes by family.

\paragraph{The settings of \sys sit on a plateau.}
Figure~\ref{fig:sens} changes one setting at a time with GPT-4o labels.
Across the sixteen tested values of $\beta_{+}$, $\beta_{-}$, $\delta$ and $\tau$, the eleven-graph mean of \sys stays between \nSensZLo\% and \nSensZHi\%, and \sysT keeps \nAllT\% for every tested $\delta$.
For both implementations one step works best.

\subsection{Generality}
\label{sec:general}

\paragraph{With open LLMs as annotators, our implementations keep the highest means.}
With Qwen3-8B labels, \sysT reaches \nQwenT\% on the eleven graphs against \nQwenSage\% for SAGE h64, the strongest baseline in Table~\ref{tab:llm}, and is significantly above the four heterophily GNNs, the best of which, H2GCN h64, reaches \nHetQwenBest\% (Table~\ref{tab:llmhet}). \sys stays above LLM-BP.
On the two additional graphs with these labels, \sysT leads SAGE h64 on all ten training seeds (Table~\ref{tab:lead}).
With Llama-3.1-8B labels, \sys reaches \nLlamaZ\% on the eleven graphs and is significantly above every baseline run with these labels, while \sysT is significantly above LLM-BP and SAGE h64, has the best mean on the two additional graphs and leads SAGE h64 on all ten training seeds (App.~\ref{app:more}).
With Mistral-7B labels, \sysT has the highest eleven-graph mean and is again significantly above every baseline.
With Qwen3.8-27B labels, \sysT reaches \nNewT\% on the eleven graphs and is significantly above every baseline in Table~\ref{tab:llm}.

\paragraph{With each of \nEncW encoder settings, both implementations stay above LLM-BP\@.}
We swap bge-en-icl with its task-aware prompt for its plain prompt, text-embedding-3-large~\cite{te3}, RoBERTa~\cite{roberta}, SBERT~\cite{sbert}, e5-base-v2~\cite{e5} and Qwen3-Embedding-0.6B~\cite{qwen3emb} one at a time, with GPT-4o labels.
\sysT has the best eleven-graph mean under \nEncTBestW of these settings, from \nEncTLo\% with SBERT to \nEncTHi\% with text-embedding-3-large, and stays at least \nEncTBPMin points above LLM-BP and above SAGE h64 under every one of them.

\begin{table}[t]
\centering
\small
\setlength{\tabcolsep}{4pt}
\caption{Eleven-graph mean accuracy in the few-shot setting of LLM-BP with $k$ true labels per class and GPT-4o labels.}
\label{tab:s2}
\begin{tabular*}{\linewidth}{@{\extracolsep{\fill}}lcccc@{}}
\toprule
Method & $k=1$ & $k=3$ & $k=5$ & $k=10$\\
\midrule
Vanilla & 46.51 & 64.12 & 68.88 & 73.61\\
Vanilla + NA & 41.78 & 57.66 & 63.30 & 68.64\\
LLM-BP & \underline{47.88} & 65.13 & 69.10 & 73.34\\
LLM-BP (appr.) & 47.70 & \underline{65.45} & \underline{69.64} & \underline{74.03}\\
\color{black!55}\sys, no LLM labels & \color{black!55}47.32 & \color{black!55}64.76 & \color{black!55}69.02 & \color{black!55}73.82\\
\sys & \textbf{67.63} & \textbf{73.66} & \textbf{74.91} & \textbf{76.16}\\
\midrule
SAGE h64 & \underline{72.25} & \underline{71.56} & \underline{72.96} & \underline{73.35}\\
\sysT & \textbf{72.39} & \textbf{73.54} & \textbf{74.02} & \textbf{74.63}\\
\bottomrule
\end{tabular*}
\end{table}

\begin{table}[t]
\centering
\small
\setlength{\tabcolsep}{2.2pt}
\caption{Accuracy on three further Web platforms with Qwen3-8B or Llama-3.1-8B as the annotator.}
\label{tab:ext}
\begin{tabular*}{\linewidth}{@{\extracolsep{\fill}}lcccccccc@{}}
\toprule
 & \multicolumn{4}{c}{Qwen3-8B} & \multicolumn{4}{c}{Llama-3.1-8B}\\
\cmidrule(lr){2-5}\cmidrule(l){6-9}
Method & Inst. & Reddit & Photo & Mean & Inst. & Reddit & Photo & Mean\\
\midrule
LLM-BP~\cite{llmbp} & \textbf{62.75} & \underline{50.50} & \underline{44.33} & \underline{52.53} & \textbf{61.67} & \underline{52.02} & \underline{24.80} & \underline{46.16}\\
\cmidrule(l){1-9}
\sys & \underline{61.05} & \textbf{54.52} & \textbf{44.82} & \textbf{53.46} & \underline{60.53} & \textbf{55.42} & \textbf{48.30} & \textbf{54.75}\\
\midrule
MLP h256~\cite{llmgnn} & \underline{42.28} & 46.57 & 42.79 & 43.88 & 44.65 & \underline{47.23} & 48.52 & 46.80\\
SAGE h256~\cite{graphsage} & 41.99 & \underline{46.72} & \underline{43.04} & \underline{43.92} & 44.80 & 47.12 & 48.93 & 46.95\\
SAGE h64~\cite{graphsage} & 41.59 & 46.11 & 42.96 & 43.56 & \underline{45.09} & 46.76 & \textbf{49.67} & \underline{47.17}\\
\cmidrule(l){1-9}
\sysT & \textbf{61.89} & \textbf{55.51} & \textbf{44.97} & \textbf{54.12} & \textbf{62.01} & \textbf{54.27} & \underline{49.46} & \textbf{55.24}\\
\bottomrule
\end{tabular*}
\end{table}

\paragraph{On three further Web platforms, \sysT has the highest mean.}
With Qwen3-8B labels on Instagram, Reddit and Photo, it reaches \nExtT\% against \nExtBP\% for LLM-BP (Table~\ref{tab:ext}).
With Llama-3.1-8B labels it again has the highest mean, and with either annotator both implementations lead every baseline of Table~\ref{tab:ext} on the mean.

\paragraph{With a few true labels per class and GPT-4o labels, both implementations stay significantly above LLM-BP\@.}
With $k$ true labels per class, \sys is significantly above few-shot LLM-BP at every $k$, and with ten labels per class \sys reaches \nFsZten\% against \nFsBPten\% (Table~\ref{tab:s2}).
\sysT is also significantly above few-shot LLM-BP at every $k$ and has the best mean with one label per class.

\subsection{Label Noise}
\label{sec:noise}

\begin{figure}[t]
\centering
\begin{subfigure}[t]{0.523\linewidth}\centering\includegraphics[width=\linewidth]{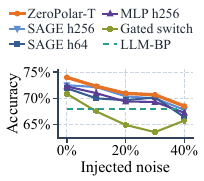}\caption{\sysT and baselines.}\end{subfigure}\hfill
\begin{subfigure}[t]{0.466\linewidth}\centering\includegraphics[width=\linewidth]{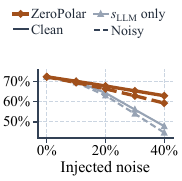}\caption{Two gates of \sys.}\end{subfigure}
\caption{Eleven-graph mean accuracy as injected noise replaces GPT-4o labels by random classes, for \sysT and baselines (a) and for two gates of \sys (b).}
\Description{Two line panels against the injected noise rate from zero to forty percent. Left, \sysT{} has the highest mean at every rate, above SAGE h256, SAGE h64, MLP h256, the gated switch and a flat LLM-BP line near 68 percent. Right, \sys{} with the larger of the two statistics declines slowly, while the gate that reads only the LLM labels falls steeply at high noise, with solid lines for clean prototype labels and dashed lines for noisy ones.}
\label{fig:noise}
\end{figure}

\paragraph{\sysT stays on top under noise.}
For noise at rate $\eta$, each remaining label of a test node is replaced with probability $\eta$ by a class drawn uniformly, with three draws per rate, while the $20C$ prototype labels stay clean except in the last variant below.
When \nNoiseMax\% of the GPT-4o labels are replaced by random classes, \sysT reaches \nNoiseT\%, and at every noise level its mean is the highest among the methods in Figure~\ref{fig:noise}a.

\paragraph{Under label noise, the prototype statistic keeps \sys far above a gate that reads only the LLM labels.}
The gate recomputes $s_{\mathrm{LLM}}$ from the perturbed labels.
At \nNoiseMax\% noise \sys reaches \nNoiseZthree\%, while a gate that reads only $s_{\mathrm{LLM}}$ falls to \nNoiseZone\% (solid lines in Figure~\ref{fig:noise}b).
By Proposition~\ref{prop:noise}, such noise can only pull $s_{\mathrm{LLM}}$ toward zero, and $s_{\mathrm{proto}}$ does not read the noisy labels.

\paragraph{Noisy prototype labels keep this order.}
When the noise also replaces the $20C$ labels behind the prototypes, the gate of \sysT keeps its sign on every graph, rate and draw, so Figure~\ref{fig:noise}a does not change.
At \nNoiseMax\% noise \sys leads the gate reading only $s_{\mathrm{LLM}}$ by \nNoiseProtoLead points, against \nNoiseLead with clean prototype labels (dashed lines in Figure~\ref{fig:noise}b).

\subsection{Cost and Cases}
\label{sec:cost}

\begin{figure}[t]
\centering
\includegraphics[width=\linewidth]{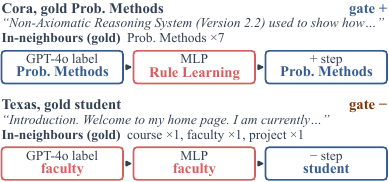}
\caption{Two nodes repaired by the signed step of \mbox{\sysT}, with the gate at $+1$ on Cora and $-1$ on Texas. Frames mark correct and wrong classes.}
\Description{Two cases stacked. Above, a Cora paper on a non-axiomatic reasoning system whose GPT-4o label is correct, whose MLP prediction is Rule Learning and whose seven in-neighbours all belong to Probabilistic Methods, and the plus step returns it to Probabilistic Methods. Below, a Texas home page that GPT-4o and the MLP label faculty, whose three in-neighbours are a course, a faculty and a project page, and the minus step moves it to student.}
\label{fig:case}
\end{figure}

\paragraph{\sys adds little to the cost of labelling each node.}
Both implementations make $20C$ more calls than SAGE h64 trained on the same labels.
On Sportsfit, our largest graph, \sys takes \nTimeZ seconds and \sysT \nTimeT seconds including the training of its MLP, against \nTimeSage seconds for SAGE h64, which also needs more GPU memory than \sysT.

\paragraph{The signed step repairs labels in both directions.}
On Cora the step toward the neighbours restores a paper that the MLP mislabels, and on Texas the step away from them turns a student page that GPT-4o and the MLP call faculty into student, while a step toward them leaves it wrong (Figure~\ref{fig:case}).
Beyond these two cases, the step of \sysT fixes more node and seed pairs than it breaks on \nFixWinsW of the eleven graphs, and on Sportsfit, the largest graph, it fixes \nFixSport pairs against \nBreakSport pairs that it breaks.

\paragraph{\sysT leads at low in-degree.}
Figure~\ref{fig:lead}b splits the test nodes into buckets of in-degree zero, one to two, three to five and at least six, and a graph enters the mean of a bucket that holds at least twenty of its test nodes.
\sysT is significantly above SAGE h64 in the three buckets with fewer than six in-edges, and above MLP h256 and LLM-BP in all four (Figure~\ref{fig:lead}b).

\section{Conclusion}
\label{sec:conclusion}
The links of a text-attributed Web graph should enter a zero-shot prediction with a sign, \polarity, on which our zero-shot node classifier \sys builds.
The excess agreement of LLM labels and of text-prototype predictions along edges reads this sign without any true label, and one signed propagation step uses it.
Training-free \sys serves small labelling budgets, and \sysT serves budgets that label at least half of the nodes by first denoising the labels with a text classifier.
With one GPT-4o label per node, \sysT reaches the highest eleven-graph mean accuracy (\nAllT\%), \nDiffBP points above LLM-BP and significantly above all \nNbaseW baselines, and with Qwen3.8-27B labels it leads the strongest baseline on two additional graphs by \nFreshDiff points.
With labels on \nBudPct\% of the nodes, \sys leads a GNN trained on as many randomly chosen labels by \nBudGap points, and on our largest graph it takes \nTimeZ seconds against \nTimeSage seconds for SAGE h64.
We expect label-free polarity checks to help other Web tasks that combine links with noisy supervision, such as link prediction and recommendation on new Web platforms.

\bibliographystyle{ACM-Reference-Format}
\bibliography{refs_gen,refs_manual}

\appendix

\section{Further Results}
\label{app:more}
\suppressfloats[t]

Table~\ref{tab:pub} lists the zero-shot accuracy that LLM-BP~\cite{llmbp} reports for six published graph models, Table~\ref{tab:freshq} the two additional graphs with Qwen3-8B labels, Table~\ref{tab:trainedfull} every trained baseline with GPT-4o labels and Table~\ref{tab:llmhet} the heterophily GNNs with open LLM labels.
Table~\ref{tab:lead} gives the lead of \sysT over the trained baselines with its patience and over ten training seeds.
App.~\ref{app:setup} gives setup details, and App.~\ref{app:locle} describes our port of Locle and its chosen configuration.

\begin{figure}[t]
\centering
\includegraphics[width=\linewidth]{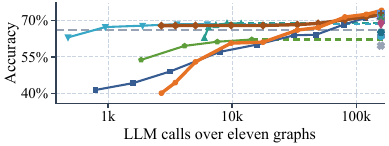}\\[1pt]
\includegraphics[width=\linewidth]{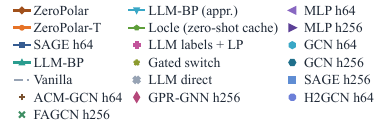}
\caption{Eleven-graph mean accuracy against LLM calls for both implementations and every baseline.}
\Description{Line plot on a logarithmic axis of LLM calls with a three-column legend below it. LLM-BP (appr.) and LLM-BP appear as short curves over their prototype budgets near 63 to 69 percent with dashed flat lines beyond their last point, and Vanilla, which makes no call, is a dashed horizontal line near 66 percent. SAGE h64 and \sysT{} climb from about 40 percent at the smallest budgets to about 72 and 74 percent at the full budget, while \sys{} stays near 68 percent and rises to about 72 percent. Locle with the zero-shot cache is a short curve over its four budgets from about 54 to 62 percent at a few thousand calls. The other baselines, including four heterophily GNNs, appear as single points at one label per test node between about 60 and 73 percent.}
\label{fig:budget_full}
\end{figure}

\begin{table*}[t]
\centering
\small
\setlength{\tabcolsep}{2pt}
\caption{Zero-shot accuracy (\%) of every trained baseline with GPT-4o labels on eleven graphs.}
\label{tab:trainedfull}
\begin{tabular}{@{}c@{\hspace{3pt}}l c ccccc cccccc ccc@{}}
\toprule
 & & & \multicolumn{5}{c}{Design graphs} & \multicolumn{6}{c}{Other graphs} & \multicolumn{3}{c}{Mean}\\
\cmidrule(lr){4-8}\cmidrule(lr){9-14}\cmidrule(l){15-17}
 & Method & Calls & Cora & Cite & Pub & Wiki & Corn & Hist & Child & Sport & Tex & Wisc & Wash & Design & Other & All\\
\midrule
\multirow{12}{*}[-2.1pt]{\rotatebox[origin=c]{90}{Trained}} & MLP h64~\cite{llmgnn} & $N$ & 75.461 & 69.512 & 90.112 & 74.055 & 66.318 & 69.137 & 35.137 & 75.227 & 84.848 & 71.195 & 65.211 & 75.091 & 66.793 & 70.565\\
 & MLP h256~\cite{llmgnn} & $N$ & 75.154 & 69.486 & 90.281 & 72.630 & 75.742 & 68.824 & 34.333 & 75.263 & \underline{87.522} & 75.094 & 70.742 & 76.658 & 68.630 & 72.279\\
 & GCN h64~\cite{gcn} & $N$ & 77.245 & \textbf{71.720} & 86.934 & 74.426 & 42.932 & 68.840 & 34.723 & \underline{78.153} & 65.241 & 58.994 & 37.991 & 70.651 & 57.324 & 63.382\\
 & GCN h256~\cite{gcn} & $N$ & 76.507 & 71.447 & 86.629 & 74.038 & 41.361 & 68.002 & 33.925 & 77.656 & 67.558 & 59.119 & 59.825 & 69.997 & 61.014 & 65.097\\
 & SAGE h256~\cite{graphsage} & $N$ & 77.122 & \underline{71.551} & 89.579 & 74.420 & 72.775 & 68.371 & 35.493 & 77.896 & 82.175 & \underline{75.220} & 72.926 & 77.089 & 68.680 & 72.502\\
 & SAGE h64~\cite{graphsage} & $N$ & 77.798 & 70.577 & 89.571 & 74.922 & 73.647 & 67.986 & \textbf{35.660} & 77.704 & 82.888 & 73.208 & 67.249 & 77.303 & 67.449 & 71.928\\
 & GPR-GNN h256~\cite{gprgnn} & $N$ & \underline{78.290} & 70.551 & 88.151 & \underline{75.891} & 60.558 & 66.855 & 35.428 & \textbf{78.351} & 72.371 & 66.038 & 64.629 & 74.688 & 63.945 & 68.828\\
 & H2GCN h64~\cite{h2gcn} & $N$ & 76.937 & 70.668 & \underline{90.357} & 74.796 & 72.077 & 68.520 & \underline{35.647} & 77.290 & 78.966 & 72.704 & \underline{73.799} & 76.967 & 67.821 & 71.978\\
 & FAGCN h256~\cite{fagcn} & $N$ & 77.122 & 70.642 & 88.937 & 74.523 & 71.030 & \underline{69.394} & 34.942 & 77.323 & 73.797 & \textbf{75.597} & 72.198 & 76.450 & 67.208 & 71.409\\
 & ACM-GCN h64~\cite{acmgcn} & $N$ & 77.860 & 70.369 & \textbf{91.176} & 74.158 & \underline{76.789} & 67.946 & 35.313 & 77.217 & \underline{87.522} & 72.579 & 72.926 & \underline{78.070} & \underline{68.917} & \underline{73.078}\\
 & Locle (zero-shot cache)~\cite{locle} & $50C$ & \textbf{78.352} & 68.342 & 82.590 & 69.534 & 44.852 & 65.483 & 27.456 & 48.795 & 69.697 & 56.855 & 42.649 & 68.734 & 51.823 & 59.510\\
\cmidrule(l){2-17}
 & \sysT & $20C{+}N$ & 77.921 & 71.070 & 90.112 & \textbf{75.953} & \textbf{78.360} & \textbf{69.775} & 35.040 & 77.336 & \textbf{88.235} & 74.717 & \textbf{75.400} & \textbf{78.683} & \textbf{70.084} & \textbf{73.993}\\
\bottomrule
\end{tabular}
\end{table*}

\begin{table*}[t]
\centering
\small
\setlength{\tabcolsep}{2.4pt}
\caption{Zero-shot accuracy (\%) of six published graph models reported by LLM-BP, next to ours with GPT-4o labels.}
\label{tab:pub}
\begin{tabular*}{\textwidth}{@{\extracolsep{\fill}}l ccccc cccccc ccc@{}}
\toprule
 & \multicolumn{5}{c}{Design graphs} & \multicolumn{6}{c}{Other graphs} & \multicolumn{3}{c}{Mean}\\
\cmidrule(lr){2-6}\cmidrule(lr){7-12}\cmidrule(l){13-15}
Method & Cora & Cite & Pub & Wiki & Corn & Hist & Child & Sport & Tex & Wisc & Wash & Design & Other & All\\
\midrule
GraphGPT~\cite{graphgpt} & 17.48 & 13.93 & 42.94 & 33.59 & 10.18 & 12.31 & 9.94 & 4.53 & 18.48 & 12.35 & 20.64 & 23.62 & 13.04 & 17.85\\
LLaGA~\cite{llaga} & 11.62 & 19.52 & 7.56 & 10.98 & 12.57 & 7.95 & 10.09 & 1.84 & 15.51 & 15.09 & 10.48 & 12.45 & 10.16 & 11.20\\
OFA~\cite{ofa} & 20.36 & 41.31 & 28.18 & 30.77 & 29.84 & 8.25 & 3.05 & 15.18 & 11.77 & 4.80 & 6.04 & 30.09 & 8.18 & 18.14\\
ZeroG~\cite{zerog} & 60.40 & 50.35 & 74.68 & 46.74 & 10.47 & 36.55 & 12.72 & 14.27 & 53.48 & 12.66 & 8.30 & 48.53 & 23.00 & 34.60\\
UniGLM~\cite{uniglm} & 45.57 & 52.26 & 70.33 & 55.05 & 23.03 & 44.24 & 21.48 & 33.46 & 21.39 & 27.16 & 24.01 & 49.25 & 28.62 & 38.00\\
GOFA~\cite{gofa} & 71.06 & 65.72 & 74.76 & 68.62 & 39.50 & 56.25 & 12.15 & 37.87 & 38.37 & 32.51 & 31.02 & 63.93 & 34.70 & 47.98\\
\midrule
\sys & 76.052 & 69.855 & 91.136 & 73.435 & 85.183 & 63.574 & 33.151 & 73.343 & 82.478 & 75.585 & 71.907 & 79.132 & 66.673 & 72.336\\
\sysT & 77.921 & 71.070 & 90.112 & 75.953 & 78.360 & 69.775 & 35.040 & 77.336 & 88.235 & 74.717 & 75.400 & 78.683 & 70.084 & 73.993\\
\bottomrule
\end{tabular*}
\end{table*}

\begin{table}[t]
\centering
\small
\setlength{\tabcolsep}{3pt}
\caption{Accuracy (\%) with Qwen3-8B labels on the two additional graphs, Computer and arXiv.}
\label{tab:freshq}
\begin{tabular*}{\linewidth}{@{}c@{\hspace{3pt}}l@{\extracolsep{\fill}\hspace{\tabcolsep}}ccc@{}}
\toprule
 & Method & Computer & arXiv & Mean\\
\midrule
 & Vanilla~\cite{bgeicl} & 33.16 & 43.93 & 38.55\\
 & LLM direct & 45.30 & 43.44 & 44.37\\
\midrule
\multirow{5}{*}[-2.1pt]{\rotatebox[origin=c]{90}{Training-free}} & LLM labels + LP~\cite{lp} & \underline{45.84} & 43.44 & 44.64\\
 & LLM-BP~\cite{llmbp} & 42.92 & 43.52 & 43.22\\
 & LLM-BP (appr.)~\cite{llmbp} & 43.79 & \textbf{44.17} & 43.98\\
 & Gated switch & \underline{45.84} & 43.93 & \underline{44.89}\\
\cmidrule(l){2-5}
 & \sys & \textbf{49.27} & \underline{44.09} & \textbf{46.68}\\
\midrule
\multirow{5}{*}[-2.1pt]{\rotatebox[origin=c]{90}{Trained}} & MLP h256~\cite{llmgnn} & 48.47 & 55.87 & 52.17\\
 & SAGE h256~\cite{graphsage} & 51.02 & 54.37 & 52.69\\
 & SAGE h64~\cite{graphsage} & \underline{53.12} & \underline{55.96} & \underline{54.54}\\
 & Locle~\cite{locle} & 48.74 & 38.72 & 43.73\\
\cmidrule(l){2-5}
 & \sysT & \textbf{57.24} & \textbf{56.21} & \textbf{56.73}\\
\bottomrule
\end{tabular*}
\end{table}

\begin{table}[t]
\centering
\small
\setlength{\tabcolsep}{1.5pt}
\caption{Mean accuracy (\%) of SAGE h64, the four heterophily GNNs and \sysT with three open LLM annotators.}
\label{tab:llmhet}
\begin{tabular*}{\linewidth}{@{\extracolsep{\fill}}l cccccccc@{}}
\toprule
 & \multicolumn{3}{c}{Qwen3-8B} & \multicolumn{3}{c}{Llama-3.1-8B} & \multicolumn{2}{c}{Mistral-7B}\\
\cmidrule(lr){2-4}\cmidrule(lr){5-7}\cmidrule(l){8-9}
Method & Other & All & Addl. & Other & All & Addl. & Other & All\\
\midrule
SAGE h64~\cite{graphsage} & 57.86 & 63.44 & 54.54 & 54.26 & 59.23 & 51.01 & 57.03 & \underline{60.63}\\
GPR-GNN h256~\cite{gprgnn} & 57.53 & 62.70 & \underline{55.16} & \underline{55.23} & 58.96 & \underline{51.45} & 56.46 & 58.91\\
H2GCN h64~\cite{h2gcn} & \underline{59.11} & \underline{64.40} & 53.09 & 55.15 & \underline{59.86} & 50.41 & 57.47 & 60.44\\
FAGCN h256~\cite{fagcn} & 58.40 & 64.19 & 54.31 & 53.98 & 58.79 & 49.62 & \underline{57.93} & 58.83\\
ACM-GCN h64~\cite{acmgcn} & 58.00 & 63.70 & 52.46 & 54.95 & 59.00 & 49.44 & 57.17 & 60.17\\
\cmidrule(l){1-9}
\sysT & \textbf{59.27} & \textbf{65.59} & \textbf{56.73} & \textbf{56.46} & \textbf{61.37} & \textbf{53.14} & \textbf{58.58} & \textbf{62.56}\\
\bottomrule
\end{tabular*}
\end{table}

\begin{table}[t]
\centering
\small
\setlength{\tabcolsep}{3pt}
\caption{\sysT minus each trained baseline with a patience of 50 on the eleven graphs with GPT-4o labels, and over ten training seeds per setting.}
\label{tab:lead}
\begin{tabular*}{\linewidth}{@{\extracolsep{\fill}}llcc@{}}
\toprule
Baseline & Setting & Difference [95\% CI] & Ahead\\
\midrule
MLP h64~\cite{llmgnn} & Patience 50 & $+3.57$ $[+2.87, +4.33]$ & \\
MLP h256~\cite{llmgnn} & Patience 50 & $+1.17$ $[+0.62, +1.76]$ & \\
GCN h64~\cite{gcn} & Patience 50 & $+7.95$ $[+6.89, +9.05]$ & \\
GCN h256~\cite{gcn} & Patience 50 & $+8.72$ $[+7.66, +9.81]$ & \\
SAGE h256~\cite{graphsage} & Patience 50 & $+0.97$ $[+0.29, +1.71]$ & \\
SAGE h64~\cite{graphsage} & Patience 50 & $+1.66$ $[+0.99, +2.35]$ & \\
H2GCN h64~\cite{h2gcn} & Patience 50 & $+1.92$ $[+1.12, +2.73]$ & \\
ACM-GCN h64~\cite{acmgcn} & Patience 50 & $+1.00$ $[+0.42, +1.59]$ & \\
\midrule
SAGE h64~\cite{graphsage} & Addl., Qwen3-8B & $+2.74$ $[+1.44, +4.03]$ & 10/10\\
 & All, GPT-4o & $+2.07$ $[+0.78, +3.34]$ & 10/10\\
 & Other, GPT-4o & $+0.24$ $[-1.15, +1.65]$ & 4/10\\
 & All, Llama & $+2.87$ $[+1.77, +3.92]$ & 10/10\\
 & Other, Llama & $+3.22$ $[+1.80, +4.71]$ & 10/10\\
 & Addl., Llama & $+2.34$ $[+1.59, +3.08]$ & 10/10\\
\midrule
SAGE h256~\cite{graphsage} & All, GPT-4o & $+0.49$ $[-0.41, +1.42]$ & 8/10\\
 & Other, GPT-4o & $-0.57$ $[-1.75, +0.60]$ & 5/10\\
\midrule
H2GCN h64~\cite{h2gcn} & All, GPT-4o & $+1.34$ $[+0.46, +2.28]$ & 8/10\\
 & Other, GPT-4o & $+0.79$ $[-0.65, +2.29]$ & 6/10\\
\midrule
ACM-GCN h64~\cite{acmgcn} & All, GPT-4o & $+0.94$ $[+0.18, +1.70]$ & 9/10\\
 & Other, GPT-4o & $+0.61$ $[-0.44, +1.65]$ & 7/10\\
\bottomrule
\end{tabular*}
\end{table}

\paragraph{Label budget.}
Figure~\ref{fig:budget_full} draws the curve or point of every method behind both panels of Figure~\ref{fig:budget}, with the baseline configurations of \S\ref{sec:setup}, Locle (zero-shot cache) at its four budgets and Vanilla, which makes no LLM call, as a dashed line.

\paragraph{Training seeds.}
Table~\ref{tab:lead} compares \sysT with SAGE h64 over ten training seeds with a seed and node bootstrap, and on the two additional graphs with Qwen3-8B labels the lead is \nSeedFresh points.
Its last three blocks retrain SAGE h256, H2GCN h64 and ACM-GCN h64 on the same ten seeds with the patience of \sysT.
On all eleven graphs, \sysT stays significantly above H2GCN h64 and ACM-GCN h64.
With Llama-3.1-8B labels, \sysT is ahead of SAGE h64 on every one of the ten seeds on all eleven graphs, the other six and the two additional graphs.

\paragraph{Settings.}
With Qwen3-8B labels, every $\delta$ from \nDeltaQLo up to the default gives the same sign on the eleven graphs and the two additional graphs, so the accuracy of \sys does not change.

\paragraph{Few-shot labels.}
With Qwen3-8B labels in the few-shot setting, \sys is again significantly above few-shot LLM-BP at every $k$.
Only with ten true labels per class does \sysT fall below few-shot LLM-BP, with \nSqTten\% against \nSqBPten\%, since \sysT learns mostly from the LLM labels and changes little with $k$, while few-shot LLM-BP rises with $k$.

\section{Setup Details}
\label{app:setup}

\paragraph{Annotators.}
Qwen3-8B~\cite{qwen3} labels all graphs as a second annotator, with at most 512 new tokens and the first class name in its answer taken as the label.
Llama-3.1-8B~\cite{llama3} labels all graphs as a third annotator with the same prompt.
It often restates the class list before answering, so we delete the restated list before taking the first class name.
On the further platforms we also match class names in the singular.
Mistral-7B~\cite{mistral}, in its instruction-tuned v0.3 release, labels the eleven graphs as a fourth annotator with the same prompt and parser as Qwen3-8B.
Qwen3.8-27B~\cite{qwen38} labels all graphs as a fifth annotator with the same prompt and parser as Qwen3-8B and with thinking turned off.
The two additional graphs carry Qwen3.8-27B labels in the main text and Qwen3-8B labels in Table~\ref{tab:freshq} of App.~\ref{app:more}.

\paragraph{Baselines.}
The sign that the released code of LLM-BP (appr.) fixes is negative on the four WebKB graphs.
The heterophily GNNs keep the default values of their own hyperparameters, and their hidden width, 64 or 256, is chosen on the design graphs.
On the two additional graphs the trained baselines are the three LLM-GNN variants with the highest design-graph means, SAGE h64, SAGE h256 and MLP h256, and with Qwen3.8-27B labels they also include the four heterophily GNNs.
GOFA-T, whose tuning data include Arxiv and Pubmed, runs with its released inference settings on all test nodes of Cora and the four WebKB graphs and on 2,000 sampled test nodes of each other graph, an unparsed answer counts as wrong, and its paired differences use the same nodes.
The Vanilla + NA row of Table~\ref{tab:s2} comes from the released code of LLM-BP.
Its neighbourhood aggregation (NA) adds the mean in-neighbour embedding of a node to its own embedding before matching this sum to the class embeddings.

\paragraph{Label budgets, noise and few-shot labels.}
For a budget $\rho<1$, a random share $\rho$ of the test nodes keeps its LLM label, with three draws per budget, so SAGE h64 then trains on the labels of randomly chosen test nodes rather than on nodes chosen by the active selection of LLM-GNN.
Each baseline configuration is one point, and its accuracy carries over to larger budgets.
Figures~\ref{fig:budget} and~\ref{fig:budget_full} count LLM calls over the eleven graphs, $20C+\rho N$ for both implementations, $\rho N$ for SAGE h64, $kC+500$ for LLM-BP, $kC$ for LLM-BP (appr.), 0 for Vanilla, $20C+N$ for the gated switch, the calls that Locle actually makes at $B\in\{20C,50C,100C,200C\}$ and $N$ for the other baselines, with $k\in\{5,10,20,40,80\}$ prototype labels per class or the largest feasible $k$ on a small graph.
For each noise draw, the trained methods share one perturbed copy, and the training-free methods share one copy per prototype seed.
In the few-shot setting of LLM-BP, $k$ true labels per class, sampled with seeds 42 to 51, replace the prototype labels and the LLM labels of the sampled nodes, and the trained rows use the first three of these seeds.

\paragraph{Seeds and training.}
Training-free rows use seeds 42 to 71 of LLM-BP with its sampling of the $20C$ prototype labels, and trained rows use training seeds 0 to 2.
The MLP of \sysT has 256 hidden units and dropout 0.5~\cite{dropout} and trains with Adam~\cite{adam} at learning rate 0.01 and weight decay $5\times10^{-4}$ for at most 300 epochs, stopping after 50 epochs without gain on a held-out tenth of the LLM labels.
The MLP, GCN, SAGE and heterophily GNN baselines follow the same recipe with a patience of 30 epochs, and all models build on the PyG library~\cite{pyg}.
On the eleven graphs, \sysT trained with the baseline patience of 30 epochs still reaches \nPatThirty\%, above every baseline.
When the MLP, GCN and SAGE baselines, H2GCN h64 and ACM-GCN h64 train with a patience of 50, \sysT stays significantly above each (Table~\ref{tab:lead}).

\section{Locle Port}
\label{app:locle}

We port the main line of the released code of Locle~\cite{locle} into one file, from its active selection of nodes through its self-training of a GCN in stages to its label correction on a rewired graph.
Its candidate pool is the test nodes with an LLM label, it uses our graphs and encoder, and it predicts every test node with the two-layer GCN of its last stage, averaged over training seeds 0 to 2.
Its configuration is one of the five in its released scripts, chosen by the mean accuracy on the five design graphs, and we remove every use of test labels, in model selection and in its entropy filter.
Its budget $B$ counts every LLM call, the first labels and the queries of later stages, and is $50C$ in Tables~\ref{tab:main} and~\ref{tab:llm} and $20C$, $100C$ or $200C$ in Figure~\ref{fig:budget_full}, at most $N$.
With GPT-4o every query takes the GPT-4o label of the node, which has no confidence, so the confidence term of its first filter and the confidence ranking of its label correction have no effect, and we name this row Locle (zero-shot cache).
With Qwen3-8B, Llama-3.1-8B and Mistral-7B it queries the LLM with its released prompt, which asks for a class and a confidence, and runs both in full (Table~\ref{tab:llm}).
The released Cora configuration, whose cluster count equals the class count, scores \nLocleAdaFo points below the chosen one with GPT-4o and \nLocleAdaQw with Qwen3-8B on the eleven graphs at $B=50C$.

\section{Use of Generative Artificial Intelligence}
\label{app:genai}
Generative artificial intelligence assistants helped draft and edit the text and write the experiment and plotting code.
An image-generation model sketched candidate layouts for Figure~\ref{fig:overview}, which we redrew as a vector diagram, and drew some of the icons in our schematic figures.
The authors checked every statement, number and figure against the experiment records and take full responsibility for the paper.

\clearpage
\end{document}